\documentclass{article}
\usepackage{ijcai25}
\usepackage{times}
\usepackage{soul}
\usepackage{url}
\usepackage[hidelinks]{hyperref}
\usepackage[utf8]{inputenc}
\usepackage[small]{caption}
\usepackage{graphicx}
\usepackage{amsmath}
\usepackage{amsthm}
\usepackage{booktabs}
\usepackage{algorithm}
\usepackage[switch]{lineno}
\usepackage{xcolor}
\usepackage{dsfont}
\usepackage{amsfonts}
\usepackage{amssymb}
\usepackage{graphicx}
\usepackage{subcaption}
\usepackage{algpseudocode}
\usepackage{tikz}
\usepackage{xcolor}
\title{Leveraging Pretrained Diffusion Models for Zero-Shot Part Assembly}

\usepackage{authblk}
\usepackage{footmisc}

\author[1]{Ruiyuan Zhang\textsuperscript{*}}
\author[2]{Qi Wang\textsuperscript{*}}
\author[1]{Jiaxiang Liu}
\author[1]{Yu Zhang}
\author[1]{Yuchi Huo}
\author[1]{Chao Wu\textsuperscript{\dag}}

\affil[1]{Zhejiang University, Hangzhou, China\\
\texttt{\{zhangruiyuan, zjljx, eehyc0, chao.wu\}@zju.edu.cn}}

\affil[2]{North China Electric Power University, Beijing, China\\
\texttt{qiwang@ncepu.edu.cn}}

\begin{document}

    \maketitle

    \begingroup
        \renewcommand\thefootnote{}
        \footnotetext{\textsuperscript{*} These authors contributed equally to this work.}
        \footnotetext{\textsuperscript{\dag} Corresponding author}
    \endgroup

    \begin{abstract}
        3D part assembly aims to understand part relationships and predict their 6-DoF poses to construct realistic 3D shapes, addressing the growing demand for autonomous assembly, which is crucial for robots. 
        Existing methods mainly estimate the transformation of each part by training neural networks under supervision, which requires a substantial quantity of manually labeled data.
        % Existing methods often rely on manual annotation of each part’s rotation and translation for novel shapes. 
        However, the high cost of data collection and the immense variability of real-world shapes and parts make traditional methods impractical for large-scale applications.
        In this paper, we propose first a zero-shot part assembly method that utilizes pre-trained point cloud diffusion models as discriminators in the assembly process, guiding the manipulation of parts to form realistic shapes. Specifically, we theoretically demonstrate that utilizing a diffusion model for zero-shot part assembly can be transformed into an Iterative Closest Point (ICP) process. Then, we propose a novel pushing-away strategy to address the overlap parts, thereby further enhancing the robustness of the method.
        To verify our work, we conduct extensive experiments and quantitative comparisons to several strong baseline methods, demonstrating the effectiveness of the proposed approach, which even surpasses the supervised learning method.
        The code has been released on \url{https://github.com/Ruiyuan-Zhang/Zero-Shot-Assembly}.
        
    \end{abstract}
    
\section{Introduction}
    3D part assembly autonomously assembles unordered 3D pieces into a realistic, complete object by predicting the rotations and translations of each piece. This research topic has drawn great attention in the field of robots in recent years, as it plays a crucial role in advancing robotic manipulation and automation~\cite{chervinskii2023auto,ghasemipour2022blocks,zhan2020generative,zhang20223d,gao2024generative}. 

    3D part assembly is challenging because of the intricate geometries and various possible assembly combinations. 
    The existing approach to 3D part assembly relies on training machine learning models with extensive manually annotated data, including rotations and scalings. However, the high cost of data collection makes it impractical to create datasets for each task, limiting supervised methods to well-resourced domains like common datasets. 
    This question drove us to search for new methods to reduce reliance on manual labeling.
    
    Diffusion models are a recent class of likelihood-based generative models that model data distributions through an iterative noising and denoising process \cite{ho2020denoising,rombach2022high}. Following this, diffusion-based distillation models \cite{wang2024prolificdreamer,liu2024one} have demonstrated significant high-fidelity 3D content generation capabilities, highlighting both their theoretical robustness and practical applicability in generating a large amount of complex 3D contents. 
    Meanwhile, some studies have also demonstrated that a pre-trained diffusion model, leveraging its density estimates, can be transferred to handle various zero-shot tasks, including classification \cite{li2023your}, semantic correspondence \cite{zhang2024tale}, segmentation \cite{tian2024diffuse}, and open-vocabulary segmentation \cite{karazija2023diffusion}. Density estimation refers to the distribution of particles in space as they evolve over time during a diffusion process.
    These works further inspire us to explore how to distill the necessary pose transformations in assembly tasks using existing diffusion models.
    
    \begin{figure*}[ht]
        \centering
        \includegraphics[width=1.0\textwidth]{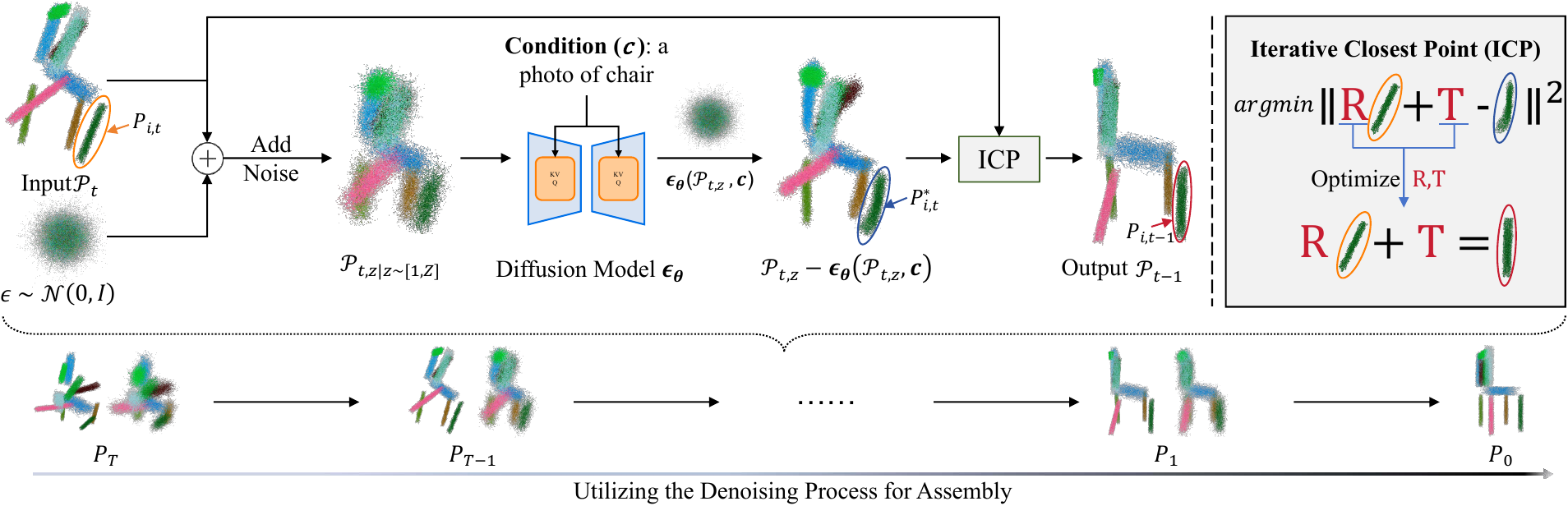}     
        \caption{
            \textbf{The overall architecture of our algorithm.}  Given the misaligned input clouds $\mathcal{P}_t$, we introduce noise to the shape, which helps the diffusion model recognize the data. The diffusion process then refines the input, generating a point cloud closer to the target chair shape. To achieve rigid transformation, we apply the ICP method for alignment, producing updated pose vectors. By iterating this process over \( T \) steps, the algorithm effectively assembles the disordered parts into the final coherent structure.
        }
        \label{fig: framework}
    \end{figure*}
    
    % 介绍我们的工作
    In this paper, we propose a new algorithm for aligning density estimates to pose transformations. 
    Specifically, we first introduce noise to a shape that has not been correctly positioned. This perturbed shape is then input into the diffusion model. The objective at this stage is to transform the disordered components into a distribution that is suitable for the diffusion model. By utilizing the denoising process, we can obtain a new point cloud that is closer to an accurate chair shape. However, it is important to note that this new point cloud does not represent a rigid transformation compared to the previous point cloud. To address this issue, we employ the ICP algorithm to align each part as closely as possible. By iteratively repeating this process, we can utilize the diffusion model to convert disordered parts into a complete shape, thereby accomplishing the entire assembly process. Since this process is performed explicitly, it allows us to apply direct pull-in or push-away operations for overlapping or distant parts, which is nearly impossible to achieve with other methods. 
    To validate our method, we employed four network architectures to predict rotational and translational transformations of parts. These baselines rely on the Shape Chamfer Distance (SCD) for supervised learning, aiming to approximate ground truth derived from reference samples generated by the diffusion model.
    Quantitative and qualitative results indicate that our method not only outperforms all baseline approaches in zero-shot settings but also surpasses some supervised techniques, underscoring its potential for practical applications.

    The contributions of our paper can be summarized as follows:
    \begin{itemize}

        \item We propose the first zero-shot assembly method that utilizes density estimates from a diffusion model to achieve continuous and smooth transformations of parts, thereby coherently assembling multiple parts. Theoretical analysis within the paper supports the efficacy of this approach.
        \item We additionally introduce a push-away strategy to mitigate collisions between parts.
        \item Results show that our method outperforms all baselines in zero-shot settings and even some supervised approaches, highlighting its practical potential.
    \end{itemize}

\section{Related Works}
    \subsection{3D Assembly Modeling}
        Estimating object pose has been a key research focus for decades. 
        In the early research, Yoon et al.~\shortcite{yoon2003real} used visual sensors and neural networks for robotic assembly. 
        Later, graph models were employed to capture semantic and geometric relationships among shape components, enabling advancements in assembly-based shape modeling \cite{zhan2020generative,jaiswal2016assembly}, while a progressive strategy leveraging the recurrent graph learning framework was explored in \cite{narayan2022rgl}. To explore the diversity of assembly outcomes, several authors propose treating parts' poses as a distribution and achieving part assembly through a diffusion process involving noising and denoising \cite{xu2024fragmentdiff,scarpellini2024diffassemble,cheng2023score}. Furthermore, innovations in network architecture have been advancing concurrently. For instance, Zhang et al.~\shortcite{zhang2024scalable} leverage the Transformer framework \cite{vaswani2017attention} to model structural relationships. Building on this, Gao et al.~\shortcite{gao2024generative} introduces hierarchical assembly to tackle the challenges associated with managing numerous parts. 
        Unlike the aforementioned works that rely on manual annotations of each part's rotation and translation, our study aims to explore a novel approach to extracting the necessary pose transformations for assembly tasks. Specifically, we investigate how existing diffusion models can be leveraged to achieve this goal, thereby reducing the dependency on labour-intensive manual labelling.

    \subsection{Diffusion Model}
        Diffusion models operate in two steps: adding noise to destroy data structure and reversing this noise to reconstruct it. This enables them to model target distributions and generate diverse content, including images \cite{saharia2022photorealistic,nichol2021glide}, videos \cite{wang2024recipe,ho2022imagen}, 3D objects \cite{peebles2023scalable,lin2023magic3d}, and audio \cite{kong2020diffwave,liu2023audioldm}.
        Recent studies suggest that diffusion models encode semantic and grouping information, leading to two main research directions. The first research direction leverages the internal representations of diffusion models for various discriminative tasks, requiring minimal additional training. These tasks include zero-shot classification \cite{li2023your}, label-efficient segmentation \cite{baranchuk2021label}, and open-vocabulary segmentation \cite{karazija2023diffusion}. The second research direction focuses on generative tasks, such as bridging 2D diffusion models and 3D generation through Score Distillation Sampling (SDS). Methods like DreamFusion \cite{poole2022dreamfusion} align 3D representations with text prompts, while later works enhance visual fidelity using strategies like coarse-to-fine optimization \cite{lin2023magic3d,chen2023fantasia3d} and multi-view consistency \cite{shi2023mvdream,hu2024efficientdreamer}. These advances highlight the versatility of diffusion models in blending discriminative and generative capabilities.
        Our work builds upon the generative approach, introducing a theoretically sound and interpretable method to tackle the zero-shot assembly problem effectively.

\section{Methodology}
    In this section, we will first provide a formal symbolic definition of diffusion models (Sec.~\ref{sec: preliminaries}). Next, we will introduce the zero-shot method proposed in this paper (Sec.~\ref{sec: zero-shot assembler}). Finally, based on our method, we will present a new approach to mitigate part overlap (Sec.~\ref{sec: collision handling}). 

        \subsection{Diffusion Model Preliminaries} \label{sec: preliminaries}
            % 介绍一下我们用到的Diffusion相关的公式就行了
            The diffusion model \cite{song2019generative,song2020score,luo2021diffusion} is a likelihood-based generative model, designed to learn the data distributions. Starting from an underlying data distribution \( q(x) \), the model applies a forward process that progressively adds noise to a data sample \( x \), creating a sequence of latent variables \( \{x_z\}_{z=1}^Z \) governed by Gaussian transition kernels $q(x_z | x_{z-1})$. At each time step \( z \), the marginal distribution of \( x_z \) is defined as:
            \begin{equation}\label{eq: forward}
                x_z \sim q(x_z | x) = \mathcal{N}(\alpha_z x, \sigma_z^2 \mathbf{I}),
            \end{equation}
            where \( \sigma_z^2 + \alpha_z^2 = 1 \), with \( \sigma_z \) gradually increasing from 0 to 1. This ensures that \( q(x_z) \) converges to a Gaussian prior distribution \( \mathcal{N}(0, \mathbf{I}) \) as \( z \) approaches \( Z \). Thus, $q(x_z)$ converges to a Gaussian prior distribution $\mathcal{N}(0, \mathbf{I})$. 
    
            The reverse process, which corresponds to the generative process, is designed to reconstruct the original data from a sequence of noisy observations. The conditional distribution $p_{\phi}(x_{z-1} | x_z)$ at each time step $z$ is modeled as a Gaussian with mean $\mu_{\phi}(x_z, z)$ and covariance variance $\Sigma_{\phi}(x_z, z)$:
            
            \begin{equation}
                 p_{\phi}(x_{z-1} | x_z) := \mathcal{N}(x_{z-1}; \mu_{\phi}(x_z, z), \Sigma_{\phi}(x_z, z))
            \end{equation}
            
            To ensure that the model can accurately reconstruct the original signal as it approaches the end of the generation process, $\Sigma_{\phi}(x_z, z)$ is typically designed to decrease as $z$ decreases. This reflects the intuition that the model's confidence in predicting the next state should increase as it gets closer to the original data point.
            
            Specifically, $\Sigma_{\phi}(x_z, z)$ can be parameterized or fixed according to a schedule that depends on the time step $z$. In practice, this variance term may be simplified to depend only on $z$, for instance, by setting it proportional to the pre-defined noise scale $\sigma_z^2$:
            
            \begin{equation} \label{eq: sampling}
                \Sigma_{\phi}(x_z, z) = \sigma_z^2 \mathbf{I}
            \end{equation}
            
            where $\sigma_z^2$ is part of a predefined noise schedule that increases over time during the forward diffusion process and consequently decreases during the reverse generative process. A linear noise schedule could be defined as:
            
            \begin{equation}
                \sigma_z^2 = \frac{\sigma_Z^2}{Z} z
            \end{equation}
            
            As such, when $z$ is small, indicating that we are close to the final generation step, $\sigma_z^2$ is also small, leading to a smaller $\Sigma_{\phi}(x_z, z)$. This design choice ensures that the model exhibits higher certainty in its predictions as it nears the reconstruction of the original data, thereby enhancing the stability and quality of the generated samples.

        \subsection{Diffusion Based Iterative Zero-Shot Assembler} \label{sec: zero-shot assembler}
            Denote the input point clouds as $ \mathcal{P} = \{P_i \mid i = 1, \dots, N \} $, where $ P_i \in \mathbb{R}^{d \times 3} $ corresponds to the $i$-th part of the 3D shape, consisting of $d$ points in the 3D space. In zero-shot task, each part of the point cloud \( P_i \) has a corresponding rigid transformation, described by a quaternion $ \text{quat}_i \in \mathbb{R}^4 $ and a translation vector $ \text{trans}_i \in \mathbb{R}^3 $, which represent the rotation and translation of the part. 
            The goal of this task is to predict the pose parameters (quaternion \( \text{quat} \) and translation vector \( \text{trans} \)) of the test samples \textbf{without} pose information during training.

            To match the current shape to the diffusion models's requirements, we introduce Gaussian noise to the current shape:
            
            \begin{equation}
            \label{eq: noise}
                \mathcal{P}_{t,z} = \mathcal{N}(\alpha_z \mathcal{P}_t, \sigma_z^2 \mathbf{I}),
            \end{equation}
            where $t$ is the iterative step of our method, $z$ is the time step in diffusion model $\epsilon$.
            
            By utilizing the denoising process, we can obtain a new point cloud that is closer to the shape's distribution:
            \begin{equation}
                \mathcal{P}_{t}^* = \mathcal{P}_{t,z} - \epsilon_\theta(\mathcal{P}_{t,z}, c), 
            \end{equation}
            where $c$ corresponds to the prompt label associated with the input sample. 
            Subsequently, to satisfy the requirements of rigid transformations, we employ ICP to obtain the vector of rotation and translation.
            We then apply the transformations to the input point cloud $\mathcal{P}_t$ to obtain the updated poses, which are then utilized as the input for the next iteration.
            
            The theoretical justification for using ICP is detailed in Section \ref{sec: theory}. By iterating the above process, we can utilize the diffusion model $\epsilon$ to convert disordered parts into a complete shape, thereby accomplishing the entire assembly process.
            
        \subsection{Collision detection and handling} \label{sec: collision handling}
            Given the explicit nature of our method, it facilitates the direct application of pull-in or push-away operations for either overlapping or distant parts.
            This strategy is very difficult to implement in existing methods due to their poses being implicitly generated by the model. 
            To describe the pushing behavior of $\mathcal{P}_i$ in a point cloud $\mathcal{P}$ to reduce overlap with $\mathcal{P}_j$ ($i \neq j$), the overlap is quantified using $\mathcal{C}(\mathcal{P}_i, \mathcal{P}_j)$, which counts coincident points. The indicator function $\mathcal{I}(\mathcal{C}(\mathcal{P}_i, \mathcal{P}_j) < \text{threshold})$ determines whether the overlap is below a predefined threshold.
            The centroids of $\mathcal{P}_i$ and their intersection region are denoted as $\bar{C}_{\mathcal{P}_i}$ and $\bar{C}_{\text{intersect}}$, respectively. The displacement required to separate $\mathcal{P}_i$ is given by:
            \[
                \Delta_{i} = \mathcal{I}(\mathcal{C}(\mathcal{P}_i, \mathcal{P}_j) < \text{threshold}) \cdot \left( \bar{C}_{\mathcal{P}_i} - \bar{C}_{\text{intersect}} \right) \cdot s.
            \]
            Here, $s$ specifies the sign of the movement. This approach computes the necessary displacement direction and distance to reduce the overlap between $\mathcal{P}_i$ and $\mathcal{P}_j$.

\section{The Theory of Zero-Shot Assembly} \label{sec: theory}
    \begin{figure*}[!ht]
        \centering
        \includegraphics[width=0.99\textwidth]{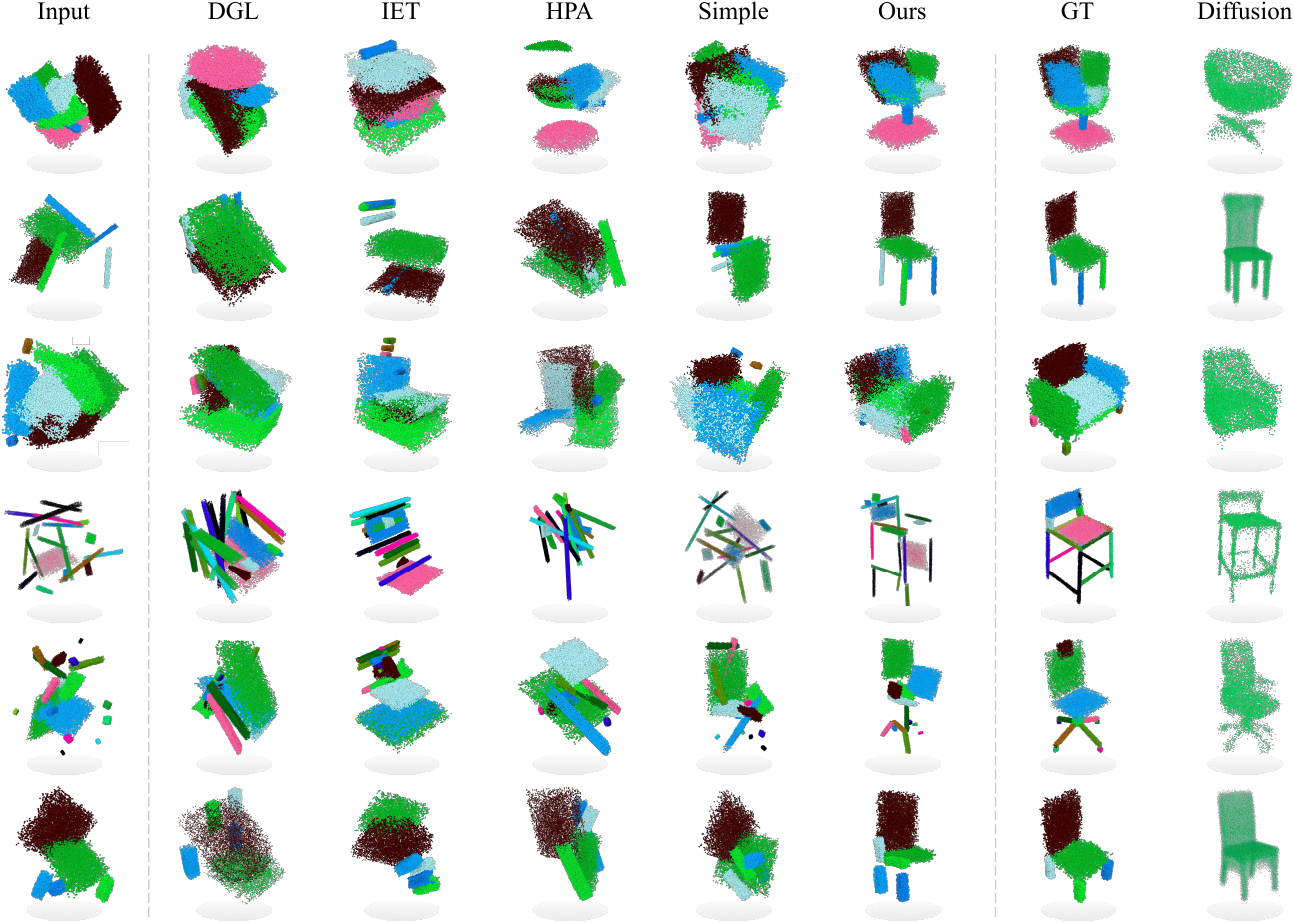}     
        \caption{
            \textbf{Visual comparisons demonstrating our superior assembly performance over baseline methods on PartNet.} 
            The first column shows our input at the Excessive level, while the last column presents reference samples obtained through diffusion sampling.
        }
        \label{fig: main visual}
    \end{figure*}
    
    \begin{figure}[tb]
        \centering 
        \includegraphics[width=0.48\textwidth]{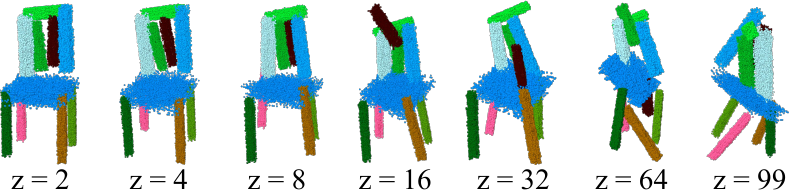}     
        \caption{
            \textbf{Different $z$ in our experiments.} 
        }
        \label{fig:different_z}
    \end{figure}

    As mentioned previously, part assembly seeks to optimize the rotation $ q_i$ and translation $t_i$ of each part $P_i$ to transform the unordered input into a coherent realistic object. Let the $q_i$ and $t_i$ be represented by an optimizable transformation matrix $\mathbf{A_i}$, then the assembly process can be formulated as:
    \begin{equation}
        \mathcal{P}_{out} = g(\mathbf{A}),
    \end{equation}
    where $\mathbf{A} = \{\mathbf{A_i} \mid i=1,...,N\}$, $g(*)$ denotes the matrix multiplication with the unordered point cloud. For previously supervised part assemblers, the optimization of $\mathbf{A}$ is quite straightforward:
    \begin{equation}
        \mathop{\arg\min}\limits_{A} \mathbb{E} \left[ g(\mathbf{A})-\mathcal{P}_{gt} \right],
    \end{equation}
    where $\mathbb{E}$ denotes a set of distance functions and $\mathcal{P}_{gt}$ is the ground truth. However, optimizing $\mathbf{A}$ is non-trivial in our case, where no supervised data is available. Therefore, instead of forcing $g(\mathbf{A})$ to fit a determined object, \textbf{we tend to make the generation of $g(\mathbf{A})$ looks like a realistic object, i.e. a sample from the distribution of the real object.} Inspired by Poole et al.~\shortcite{poole2022dreamfusion}, we leverage a pre-trained diffusion model for 3D point cloud generation, which implicitly captures the distribution of point clouds in real-world objects. Then we optimize over $\mathbf{A}$ so that $g(\mathbf{A})$ looks like a sample from this frozen diffusion model. This is achieved through a Score Distillation Sampling (SDS) loss \cite{poole2022dreamfusion}:
    \begin{equation}\label{eq: sds}
        \nabla_\mathbf{A} \mathcal{L}_{\mathrm{SDS}}(\theta, g(\mathbf{A})) \triangleq \mathbb{E}_{z, \epsilon}\left[w(z)\left({\epsilon}_\theta\left(\mathcal{P}_{t,z} ; c, z\right)-\epsilon\right) \frac{\partial \mathcal{P}_{out}}{\partial \mathbf{A}}\right].
    \end{equation}
    As shown in Fig.\ref{fig: framework}:
    \begin{align} 
        \label{eq: eps1}
        {\epsilon}_\theta\left(\mathcal{P}_{t,z} ; c, z\right) &= \mathcal{P}_{t,z}-\mathcal{P}^*, \\
        \label{eq: eps2}
        \epsilon &= \mathcal{P}_{t,z}-\mathcal{P}_{t}.
    \end{align}
    Substituting Eq. \ref{eq: eps1} and Eq. \ref{eq: eps2} into Eq. \ref{eq: sds}, we get:
    \begin{equation}\label{eq: sds_new}
        \nabla_\mathbf{A} \mathcal{L}_{\mathrm{SDS}}(\theta, g(\mathbf{A})) \triangleq \mathbb{E}_{z}\left[w(z)\left(\mathcal{P}_t-\mathcal{P}^*\right) \frac{\partial \mathcal{P}^\text{out}}{\partial \mathbf{A}}\right].
    \end{equation}
    In the equation above (Eq. \ref{eq: sds_new}), since $g(*)$ represents matrix multiplication, $\frac{\partial \mathcal{P}_{out}}{\partial \mathbf{A}}$ corresponds to the coordinates of the points in $\mathcal{P}_t$, which are constant three-dimensional vectors; $w(z)$ is a constant scalar in practice, which will be explained in the following section. Therefore, the remaining term 
    $\mathcal{P}_t-\mathcal{P}^{*}$ governs the descent process of the $\mathcal{L}_{\mathrm{SDS}}$. If we can accurately estimate the transformation from $\mathcal{P}_t$ to $\mathcal{P}^{*}$, the optimization process will converge directly.
    
    In practice, we utilize the ICP algorithm to estimate the transformation between $\mathcal{P}_t$ and $\mathcal{P}^*$, as shown in Fig. \ref{fig: framework}. It is worth noting that the smaller the change in the shape of each part in $\mathcal{P}_t$ and $\mathcal{P}^*$, the more accurate the transformation obtained by the ICP algorithm. We minimize the shape variation between $\mathcal{P}_t$ and $\mathcal{P}^*$ by controlling the magnitude of noise added and removed during the forward and generation processes. Recall that the step size determines the noise (Eq. \ref{eq: forward} and \ref{eq: sampling}): (1) In the forward process, smaller step sizes reduce the Gaussian noise variance, bringing  $\mathcal{P}_t$ closer to $\mathcal{P}_{t,z}$; (2) In the generation process, smaller step sizes reduce denoising variance, making $\mathcal{P}_{t,z}$ closer to $\mathcal{P}^*$. Therefore, we fixed the time step $z$ to a small value, typically 2 or 4, to obtain more accurate ICP estimates. Fig. \ref{fig:different_z} shows the assembly result under different $z$ with the same iterations, a smaller $z$ significantly improves the realism of the results.
    
    We finally apply the transformations obtained from the ICP algorithm to $\mathcal{P}_t$ to generate the result of this iteration $\mathcal{P}_{t-1}$ and use it as the input for the next iteration. With each iteration, the diffusion model helps bring our results closer to real-world objects.
    
\section{Experiments}
    \begin{table*}[!t]
        \centering
        \caption{
            \textbf{Quantitative evaluation on zero-shot scenario.} \underline{Underline}/\textbf{bold} fonts highlight the suboptimal/best approach. Our approach outperforms current methods in addressing the zero-shot challenge.
        }
        \begin{tabular}{llccccc}
            \toprule[1pt]
            \noalign{\vskip -1.2mm}  
            \midrule
                Methods & Noise Level & SCD $\downarrow \times 10^{-3} $ & PA $\uparrow$ \% & RMSE(Trans) $\downarrow$ $\times 10^{-2}$ & RMSE(Rot) $\downarrow$ & fPA $\uparrow$ \% \\
            \midrule 
                Ours & Slight & 7.7 & 68.59 & 7.56 & 7.18 & 68.91 \\
            \midrule 
                Ours & Moderate & 17.0 & 36.53 & 27.19 & 27.16 & 37.43 \\
            \midrule 
                Ours & Substantial & 34.8 & 15.54 & 26.89 & 32.70 & 17.90 \\
            \midrule 
            \noalign{\vskip -1.2mm}
            \midrule 
                Ours & Excessive & 45.0 & \textbf{9.0} & \textbf{28.52} & \textbf{31.02} & \textbf{12.3} \\
            \midrule 
                Simple & Excessive & \underline{31.7} & \underline{2.49} & 48.13 & \underline{57.13} & \underline{6.26} \\
            \midrule 
                HPA & Excessive & 156.2 & 0.02 & \underline{42.69} & 72.41 & 0.06 \\
            \midrule 
                IET & Excessive & \textbf{12.94} & 0.05 & 56.75 & 93.12 & 0.18 \\
            \midrule 
                DGL & Excessive & 165.2 & 0.03 & 66.80 & 69.35 & 0.08 \\
            \midrule
                Random Initial & Excessive & 203.4 & 0.0 & 61.48 & 89.97 & 0.0 \\
            \midrule
            \noalign{\vskip -1.2mm}  
            \bottomrule[1pt]
        \end{tabular}
        \label{table:main_result}
    \end{table*}
    \begin{figure}[tb]
        \includegraphics[width=0.48\textwidth]{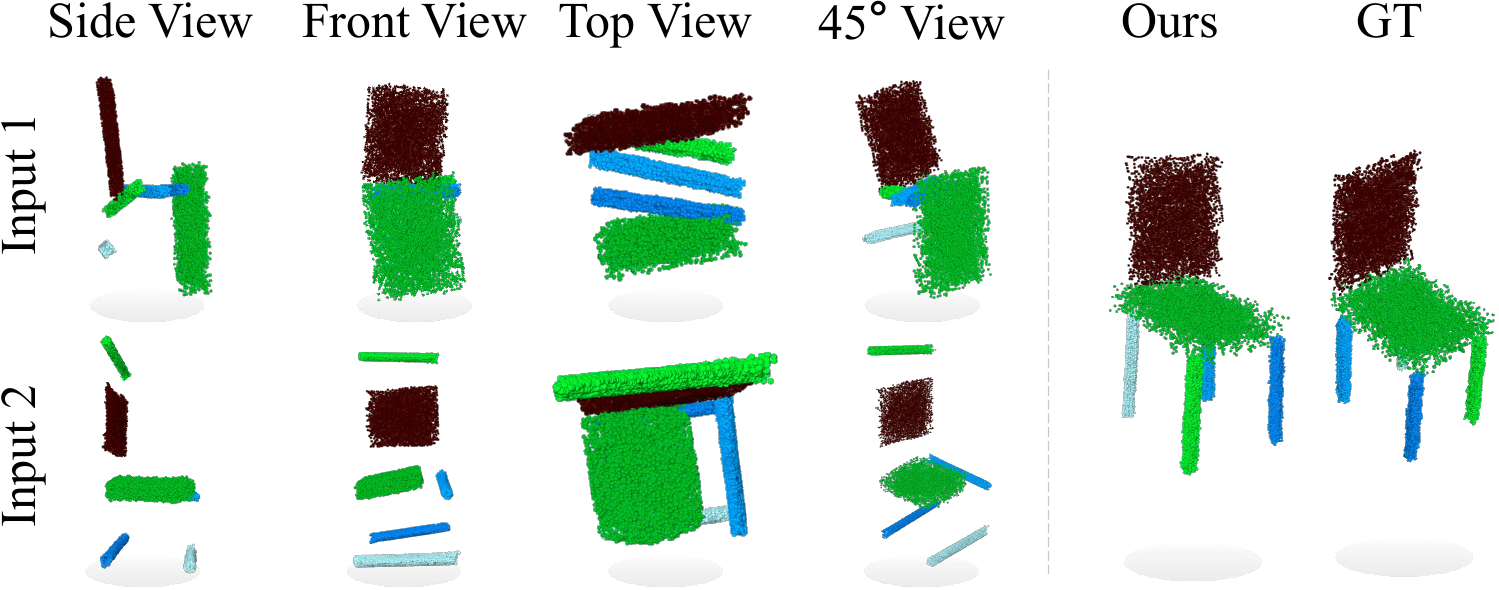}     
        \caption{
            \textbf{Different Views from Baseline-Simple and Ours.} Baseline-Simple utilizes supervised learning on point clouds generated by a diffusion model, while our method employs density estimates. The results of the Simple are similar to set of point clouds from reference, but do not correspond to a chair shape.
        }
        \label{fig: different views}
    \end{figure}
    \subsection{Datasets, Baselines, and Metrics}
\begin{figure*}[ht] % 使用 figure* 环境跨越双栏
        \centering % 居中对齐
        \begin{minipage}[b]{0.46\linewidth}
            \centering % 子图内部居中
            \includegraphics[width=\linewidth]{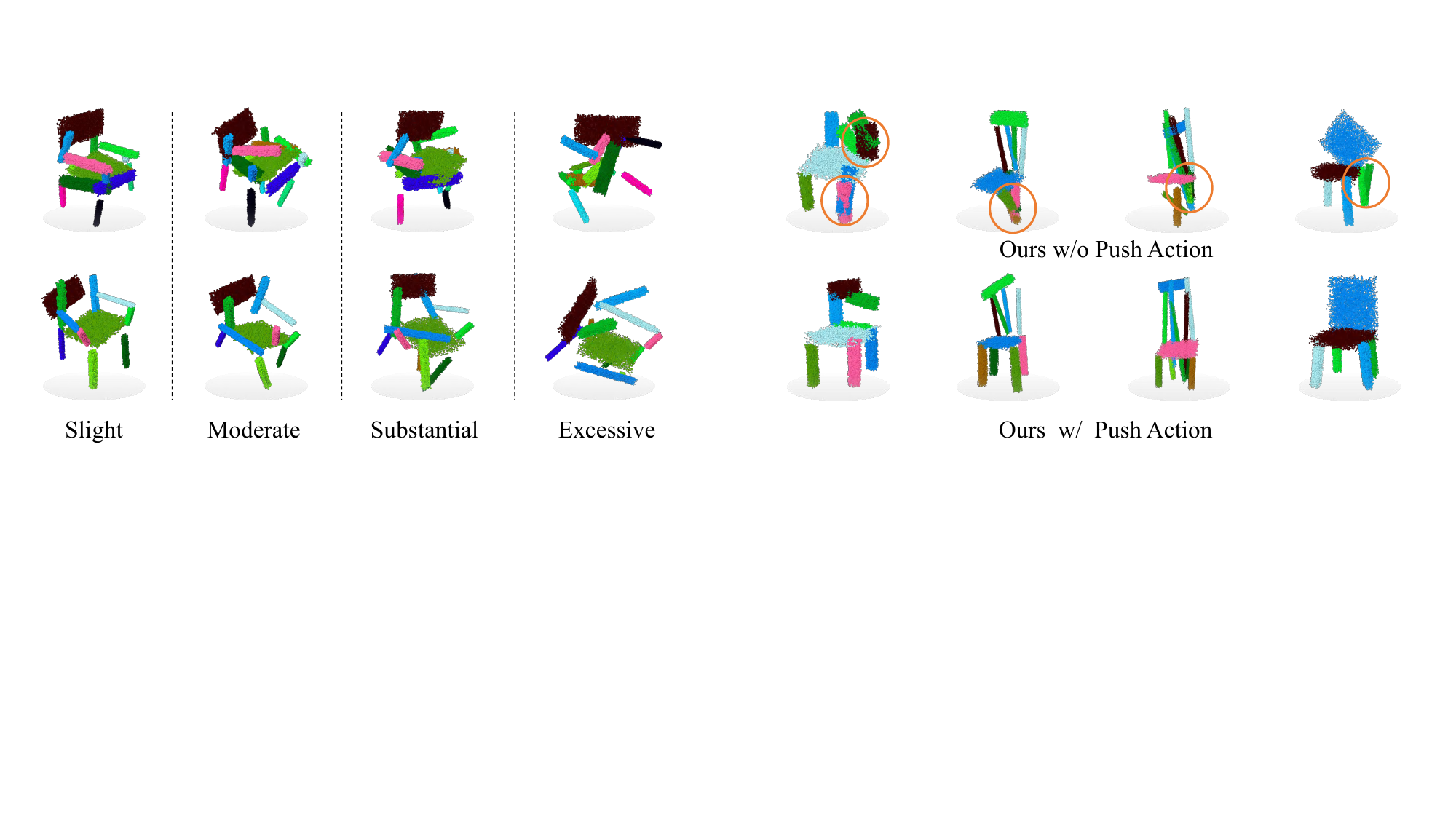}
            \caption{
                \textbf{Ablation Study of Push Action.}
                Based on our method, we can clearly separate overlapping parts, which helps reduce the overlap problem.
            } % 使用 caption* 来添加无编号的标题
            \label{fig: push_action}
        \end{minipage}
        \hfill % 两个 minipage 之间的空白
        \begin{minipage}[b]{0.46\linewidth}
            \centering % 子图内部居中
            \includegraphics[width=\linewidth]{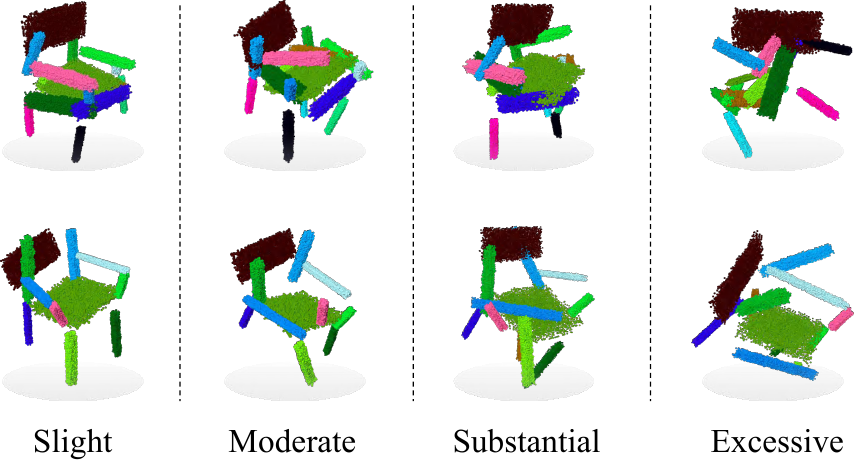}
            \caption{
                \textbf{Different noise levels in our experiments.} 
                Illustrations of input under various noise conditions, including slight, moderate, substantial, and excessive noise.
            } % 使用 caption* 来添加无编号的标题
            \label{fig: different_noise_input}
        \end{minipage}
        % \vspace{-160pt} % 可选：调整标题与图形之间的间距
    \end{figure*}

        % Baselines
        \textbf{Dataset.} We evaluate our method using assembly benchmark datasets: PartNet \shortcite{mo2019partnet}, a large-scale shape dataset with fine-grained and hierarchical part segmentations, for both training and evaluation. We use its Chair subset and adopt the dataset's default train/test/validation splits. We utilize the training set to train a diffusion probabilistic model for 3D point cloud generation, while the test set data is used for zero-shot assembly. The number of parts ranges between 2 and 20. As shown in Fig.~\ref{fig: different_noise_input}, to assess the robustness of both baseline methods and our proposed approach, we established four distinct noise levels: \textit{slight}, \textit{moderate}, \textit{substantial}, and \textit{excessive}. Finally, the 3D diffusion model in this work was come from previous work \cite{luo2021diffusion}.

        \noindent\textbf{Comparison of Baselines.} We compared our approach with Complement \cite{sung2017complementme}, DGL \cite{zhan2020generative}, IET \cite{zhang20223d}, HPA \cite{gao2024generative}, and Simple. Among them, Simple is the Baseline we designed, which utilizes seven trainable parameters to represent the rotational and translational transformations of parts. Simple usually outperforms other baseline methods in practice.
        
        \noindent\textbf{Evaluation Metrics.}
        To conduct a comprehensive evaluation, we employ a set of metrics that include Part Accuracy (PA) used in Zhan et al.~\shortcite{zhan2020generative}, along with Shape Chamfer Distance (SCD) as used by Zhan et al.~\shortcite{zhan2020generative}. Additionally, we incorporate Root Mean Squared Error for Rotation (RMSE(R)) and Translation (RMSE(T)) as outlined in Sell{\'a}n et al.~\shortcite{sellan2022breaking}. Specifically, PA measures the precision of each part, SCD evaluates the overall shape quality, and RMSE(R) and RMSE(T) gauge the accuracy of rotation and translation predictions.

        \noindent\textbf{Fair Part Accuracy (fPA).} Vanilla PA is determined by the Chamfer Distance between components with identical tensor indices. This metric serves as a criterion for evaluating assembly precision. However, certain components, such as stool legs, are permitted to be positioned in regions with inconsistent indices, as illustrated by GT and Ours in Figure. Therefore, we propose the concept of Fair Part Accuracy (fPA).
        Given two point clouds $\mathcal{P}_{\text{pred}} = \{p[i], i \in \{1, 2, \ldots, N\}\}$ and $\mathcal{P}_{\text{gt}} = \{g[i], i \in \{1, 2, \ldots, N\}\}$,  $\mathcal{CD}$ represents Chamfer Distance calculation. we formally define:

        \[
            j^* = \operatorname{argmin}_{j}  \mathcal{CD}(\mathcal{P}_{\text{pred}}[i], \mathcal{P}_{\text{gt}}[j]).
        \]

        \noindent Next, we use $\mathcal{P}_{\text{fair\ gt}}$ to replace $\mathcal{P}_{\text{gt}}$.
        \[ 
            \mathcal{P}_{\text{fair\ gt}}[i] = \mathcal{P}_{\text{gt}}[j^*], \text{ where } i \in \{ 1, 2, \ldots, N \}. 
        \]
        
        \noindent We define the accuracy as:
        \[
            \text{fPA} = \frac{1}{N} \sum_{p=1}^{N} \mathds{1} \left( \frac{1}{N} \sum_{p} \mathcal{CD}(\mathcal{P}_{pred}[p], \mathcal{P}_{fair~gt}[p]) < \text{thre} \right),    
        \]
        where \text{thre} = 0.01, which is a parameter inherited from previous work \shortcite{zhan2020generative}. $\mathds{1}$ denotes an indicator function that equals 1 if the condition inside is met, and 0 otherwise.

    \subsection{Experiments Results and Analysis}
        As demonstrated in the first four rows of Table~\ref{table:main_result}, assembly performance declines with increasing noise intensity, thereby validating our noise level designations.
        The \textit{slight} noise level evaluates the ability of our method to converge close to the ground truth. Conversely, the \textit{excessive} noise level, characterized by randomly dispersed point clouds, tests the extremes of performance for both baselines and our method. The intermediate \textit{moderate} and \textit{substantial} levels further substantiate the efficacy and rationality of our noise addition strategy. 

\begin{figure*}[t]
        \centering
        \includegraphics[width=0.8\textwidth]{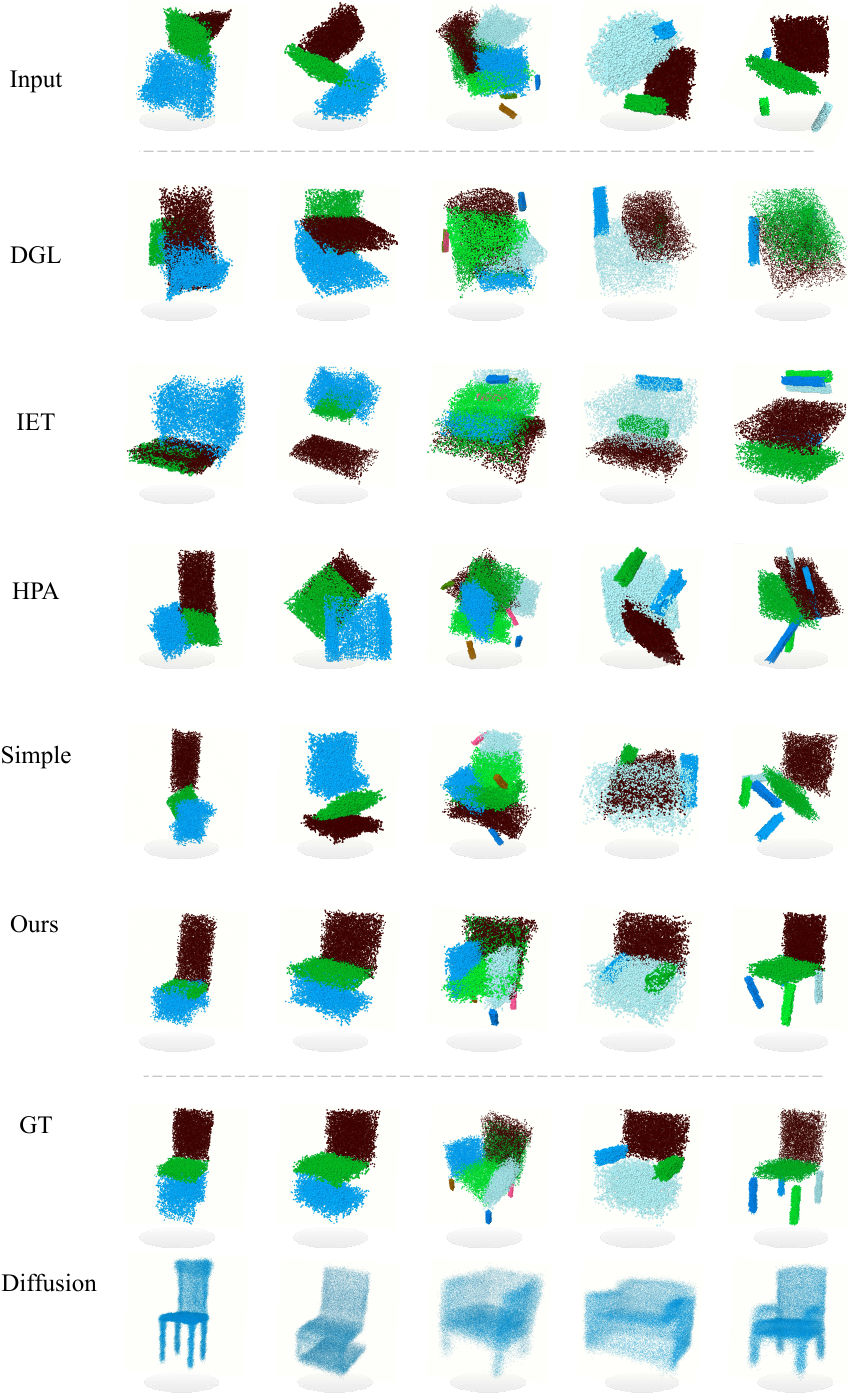}     
        \caption{
            \textbf{More visual comparisons on PartNet.} 
        }
        \label{fig: appendix; more visual}
    \end{figure*}

        We evaluated our method against various baselines, as shown in Table~\ref{table:main_result}, Fig.~\ref{fig: main visual}, and Appendix Fig.~\ref{fig: appendix; more visual}. Our approach outperforms current methods in addressing the zero-shot challenge. 
        All of our metrics outperform existing methods, except for SCD.
        % 对这个结果的解释
        This is expected since our method emphasizes density estimates from a diffusion model rather than sampling complete shapes. Therefore, our assembled outputs conceptually resemble chairs instead of precisely replicating chair-shaped point clouds. This distinction is illustrated in Fig.~\ref{fig: different views}, which visualizes samples generated by Simple under two different random seeds.
        % 对实验结果的进一步分析
        Furthermore, compared to IET, Simple performs worse on the SCD, highlighting the advantage of Transformer models in encoding complex structures. Additionally, we tested a carefully designed model, HPA, whose performance is significantly impaired when trained exclusively with the SCD. Without any prior information on part poses, all baselines demonstrate notably poor performance, as further analysis of their training loss functions reveals why they fail in zero-shot scenarios (details in Appendix Section~\ref{appendix sec: baseline fail}). Appendix Section~\ref{appendix sec: failure case} presents an experiment designed to illustrate both the effectiveness of the proposed method and its limitations on challenging samples.
        % 更多的可视化结果在Appendix Sec~\ref{fig: main visual}

\begin{figure*}[t]
            \centering
            \includegraphics[width=0.93\textwidth]{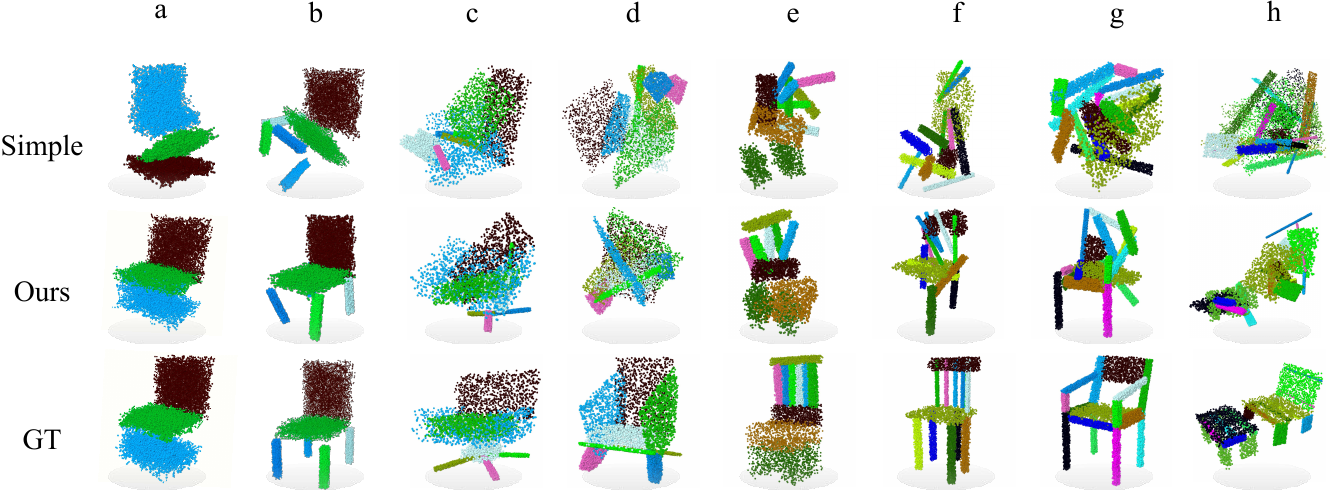}
            \caption{Performance of our method in assembly scenarios with different levels of complexity.}
            \label{fig:zero-shot-performance-under-different-complexities} 
        \end{figure*}

        To evaluate the proposed method under varying levels of task complexity, we conducted experiments with different numbers of components. As shown in Fig.~\ref{fig:zero-shot-performance-under-different-complexities}, assembly difficulty increases notably with the number of components. Compared to Simple, our method consistently achieves superior performance across all levels of complexity. Fig.~\ref{fig:zero-shot-performance-under-different-complexities}(a–h) further reveals a performance decline in zero-shot settings as the component count increases. Notably, when the number exceeds 10, assembly errors become more prominent. Additionally, as illustrated in Fig.~\ref{fig:zero-shot-performance-under-different-complexities}(c, d, e), the presence of large-volume components exacerbates task difficulty, demanding greater robustness from the method.

    \subsection{Comparisions with supervised scenario}
        \begin{table}[ht]
            % \small
            \centering
            \caption{
                \textbf{Comparisions with methods on Supervised scenario.} 
                PA is a metric used to evaluate the accuracy of each part. Our zero-shot method can surpass the Complement with supervised learning.
            }
            \begin{tabular}{llccc}
                \toprule
                    Scenario & Methods & SCD $\downarrow $ & PA $\uparrow$ & fPA $\uparrow$ \\
                \midrule 
                    Zero-Shot & Ours & 45.0 & \underline{9.0} & 12.3 \\
                \midrule 
                    Zero-Shot & Simple & 31.7 & 2.49 & 6.26 \\
                \midrule
                    Supervised & Complement & \underline{24.1} & 8.78 & - \\
                \midrule
                    Supervised & DGL & \textbf{9.1} & \textbf{39.0} & - \\
                \bottomrule[1pt]
            \end{tabular}
            \label{table: supervised scenario}
        \end{table}
        As indicated in Table \ref{table: supervised scenario}, our work achieves comparable results by the early supervised learning method: Complement. 
        This finding underscores that our method can deliver competitive outcomes even compared to pose-accessible supervised learning. 
        While our work may not yet achieve the performance of existing well-designed supervised learning methods, we hope it offers insights that may contribute to future research in zero-shot learning.
        
    \begin{figure*}[th]
        \centering
        \includegraphics[width=1.0\textwidth]{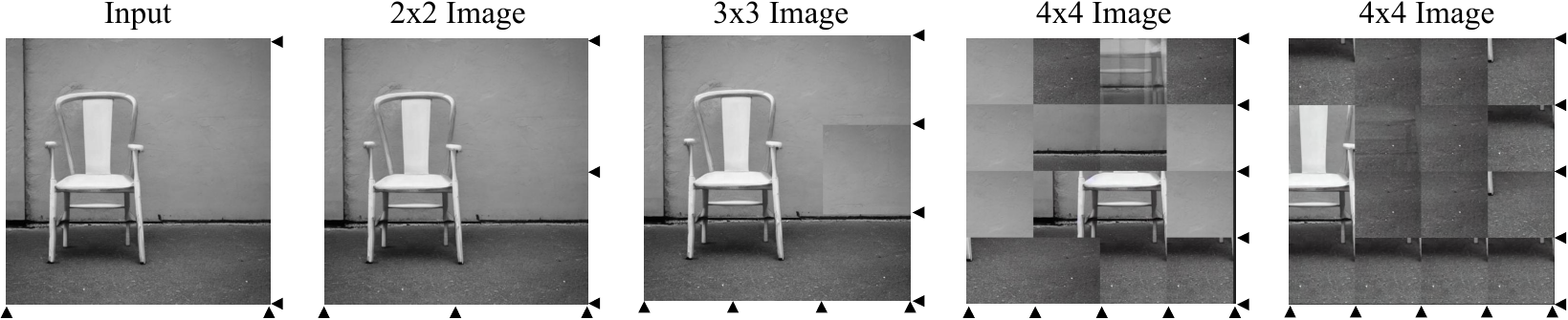} 
        \caption{
            \textbf{Visual results of 2D Image Reassembly.} This figure showcases the effectiveness of the 2D diffusion model in reassembling fragmented images without simplifying the problem, achieving near-perfect reconstruction for 3$\times$3 image puzzles. 
        }
        \label{fig: 2d reassembly}
    \end{figure*}
    \begin{figure}[ht]
        % \centering
        \includegraphics[width=0.48\textwidth]{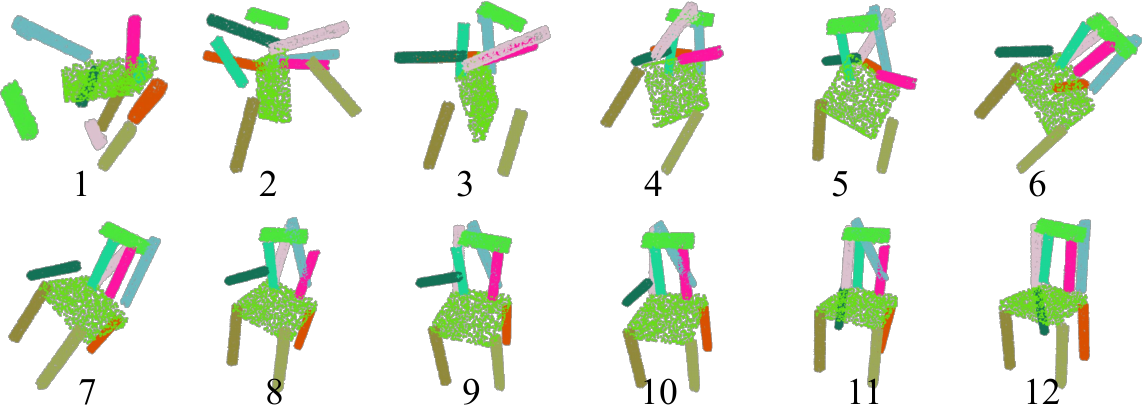} 
        \caption{
            \textbf{The process of step-by-step assembly with 2D Stable Diffusion.} 
            By simplifying the experiments' complexity, the 3D part assembly can also be achieved using 2D stable diffusion.
        }
        \label{fig: toy experiments}
    \end{figure}

    \subsection{Analyzing Effectiveness and Failure Cases from a Diffusion Perspective} \label{appendix sec: failure case}

        \begin{figure}[ht]
            % \centering
            \includegraphics[width=0.48\textwidth]{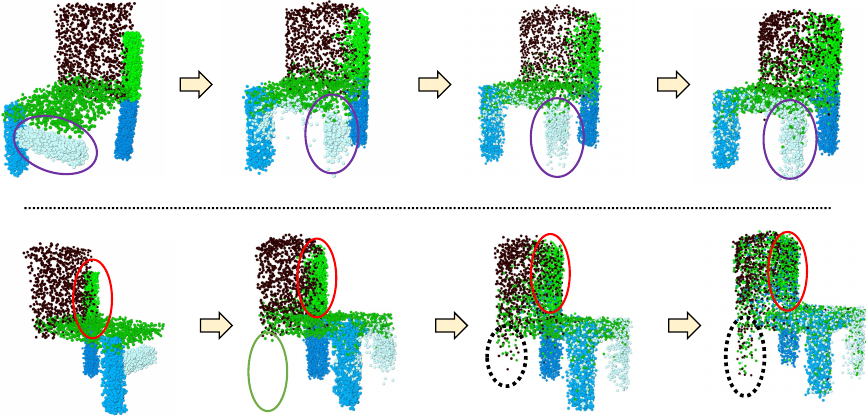} 
            \caption{Analyzing effectiveness and failure cases from a Diffusion perspective.}
            \label{fig: failure case}
        \end{figure}

        This subsection presents an experiment designed to illustrate both the effectiveness of the proposed method and its limitations on challenging samples. As shown in the last row (Ours) of Figure~\ref{fig: main visual}, the assembled shape is incorrect: the cyan chair legs are misplaced at the same height as the chair back. We refer to this issue as severe misalignment, typically caused by the uncontrollable randomness in the 3D part initialization process. In such cases, the pretrained diffusion model, under a zero-shot setting, can only partially infer the correspondence between point cloud segments and part semantics, limiting its ability to resolve these errors.
        
        To better understand this behavior, we design an experiment that analyzes the method from the diffusion process perspective. Specifically, we take an incorrectly assembled shape with severe misalignment as input and apply the point cloud diffusion model for denoising. During this process, points within parts are allowed to move freely, not constrained to rigid transformations.
        As shown in the upper part of Figure~\ref{fig: failure case}, the part in the region marked by the purple circle \tikz[baseline=-0.75ex]\draw[color={rgb,255:red,128; green,0; blue,128}, thick] (0,0) circle (0.15cm); is guided to a more reasonable position. This transformation resembles the effect of an ICP-like alignment, explaining why the proposed method often succeeds.
        
        However, in the lower part of the same figure, the part marked by the red circle \tikz[baseline=-0.75ex]\draw[red, thick] (0,0) circle (0.15cm); remains significantly misaligned from its intended location (green circle \tikz[baseline=-0.75ex]\draw[green, thick] (0,0) circle (0.15cm);). To complete the global shape, the diffusion model blends the red part’s points into the chair back, and concurrently shifts some points from the back toward the green region, forming a mixed segment at the position marked by the black dashed circle \tikz[baseline=-0.75ex]\draw[black, thick, dashed] (0,0) circle (0.15cm);. Such behavior, involving point-level redistribution rather than rigid transformation, is beyond the capacity of ICP-like processes, thus revealing why the proposed method struggles with severe misalignments. By clearly identifying this limitation, we aim to encourage future work toward more robust zero-shot assembly solutions.
        
    \subsection{Ablation Study}
        In this work, we propose an improved approach to address the collision issue that arises when identical parts are placed in the same position. To mitigate this issue, we introduce an explicit pushing-apart operation. Traditional model-based training methods struggle to achieve this directly, as the model generates the predicted poses of parts, and adjustments can only be made by tuning the model parameters, which limits operational flexibility.  
        However, in this study, we innovatively incorporate an explicit pushing-apart operation into the original method. This operation effectively separates parts, thereby reducing collision issues. The experimental results, as shown in Fig.~\ref{fig: push_action}, demonstrate the effectiveness of the proposed method.

    \subsection{Application to Zero-Shot Airplane Assembly}

        The proposed method is not limited to zero-shot chair assembly; this section further evaluates its effectiveness on airplane models. A point cloud diffusion model pretrained on the airplane category of ShapeNet \cite{chang2015shapenet} serves as the generative module. Qualitative examples are randomly selected from the test set. Airplane models are manually segmented into wings, fuselage, and tail, then perturbed with excessive noise before assembly using our method.
        
\begin{figure}[ht]
            % \centering
            \includegraphics[width=0.48\textwidth]{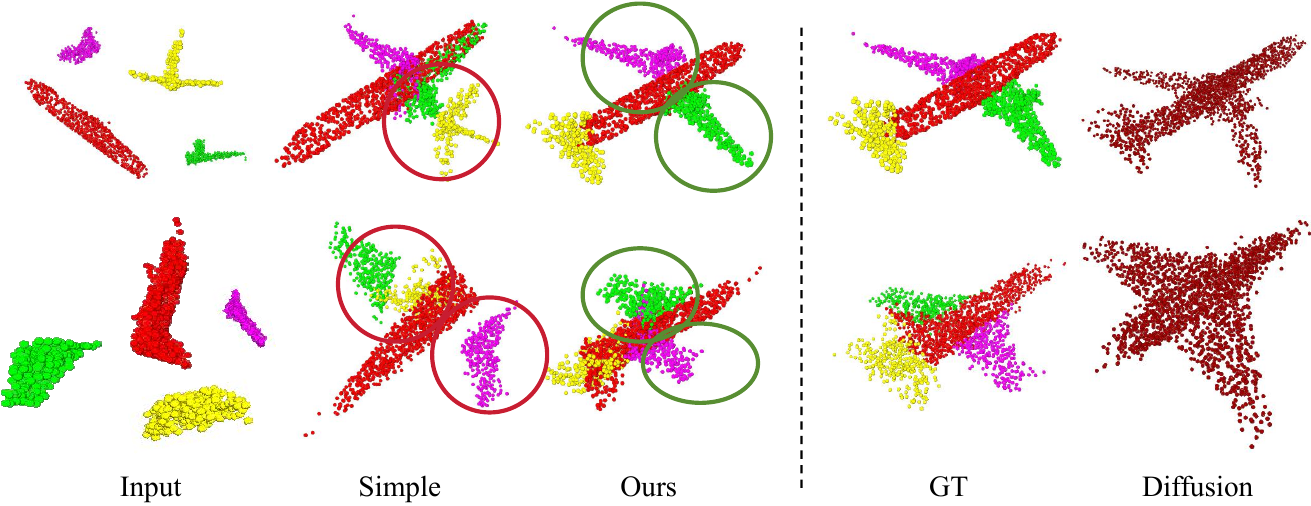} 
            \caption{Visual results of zero-shot airplane assembly.}
            \label{fig: airplane}
        \end{figure}

        As shown in Figure~\ref{fig: airplane}, the method successfully assembles disordered components into coherent airplane structures. In contrast, the baseline Simple tends to force component features to match those of a reference sample, often resulting in incorrect assemblies (e.g., red circles). Our method yields more plausible results (e.g., green circles) due to its integration of a collision detection and resolution module, which effectively reduces component overlap. These results further demonstrate the generalizability and robustness of the proposed approach across diverse 3D assembly tasks.

    \subsection{Tranfer our work to 2D Diffusion Model}
        As analyzed in Section~\ref{sec: theory}, our method is not limited to specific types of diffusion models. A pretrained 2D image diffusion model, in theory, can also evaluate the quality of assembly results by differentiably rendering 3D components into 2D space and feeding them into the model. However, in practice, this approach is challenging due to the difficulty of propagating 2D signals back to 3D point clouds and pose parameters. To test this potential, we conducted a simplified experiment where all parts shared a common rotational perturbation. This perturbation was optimized using Eq.~\ref{eq: sds}. We utilized stable diffusion 2.1 \cite{Rombach_2022_CVPR} with the prompt "a picture of colorful chair" to generate a realistic chair shape. As shown in Fig.~\ref{fig: toy experiments}, we demonstrate the visualization of the assembly process. These images are processed through microrendering, serving as inputs for the diffusion model to obtain SDS loss. This experiment demonstrates that 2D diffusion models have inherent assembly potential, though sophisticated methods are required to fully utilize it. More details of the experiment will be provided in Appendix Section~\ref{appendix sec: 3d assembly with 2d diffusion}.

    \subsection{Transfer our work to 2D Image Reassembly}
        In addition to exploring the potential of 2D diffusion models for 3D part assembly tasks, we also examined their capability in 2D image reassembly. Our model can almost perfectly reconstruct 3×3 image fragments without oversimplifying the challenge. 2D Image Reassembly involves reassembling cropped segments of a 2D image \cite{scarpellini2024diffassemble}. In this experiment, no special architecture was designed; instead, we implemented a classifier utilizing two CNN layers. An MLP was implemented to predict the correct position of each sub-image within the whole picture. Training followed the method outlined in Eq.~\ref{eq: sds}, using stable diffusion 2.1 \cite{Rombach_2022_CVPR} with the prompt "a picture of chair". As illustrated in Fig.~\ref{fig: 2d reassembly}, our approach successfully handles 2D image reassembly challenges up to a complexity of 4×4. More details of the experiment will be provided in Appendix Section~\ref{appendix sec: 2d assembly}.

\section{Conclusion}
    In this work, we introduced a novel zero-shot assembly method that leverages the inherent assembly capabilities of general-purpose diffusion models to generate continuous rigid transformations for object assembly without prior training on specific shapes or configurations. It uncovers the implicit assembly abilities of general models, enabling assembly tasks even with previously unseen data. 
    Although there is a substantial amount of research in the field of diffusion models, there is relatively little research on how to apply these models to assembly tasks.
    Our approach aims to harness existing extensive work on diffusion models to achieve assembly at virtually no additional cost, which represents a meaningful and valuable contribution. 

\section{Limitation and Future Work}
    % fail case
    The bottom row of Fig.~\ref{fig: main visual} exemplifies a failure case, revealing challenges in accurately placing overlapping parts. Especially when they are far from their GT positions, using the push operation cannot accurately place overlapping parts. Adjusting random inputs can mitigate this issue, but it is not robust enough. 
    In the future, we plan to investigate an interpretable approach to reposition misplaced parts. This requires us better to explore the underutilized assembly knowledge inherent in general models and to be able to identify which positions have vacancies, thus allowing for the effective transfer of parts.
    Our goal is to improve the model’s performance in assembly tasks without extensive additional costs or extensive supervised learning.

% \clearpage
\section{Appendix}
    % \vspace{360pt}
        
    \subsection{Zero shot 2D Assembly with 2D Diffusion Model} \label{appendix sec: 2d assembly}
        In this experiment, we designed a model consisting of a two-layer Convolutional Neural Network (CNN) and a Multi-Layer Perceptron (MLP). We use trainable embedding as input to represent each sub-image. We used the softmax activation function to predict which sub-image should be chosen for each location, and we applied the SDS loss function to optimize the model. Specifically, we take the predicted values (logits) for each sub-image, multiply them by the sub-image pixels, and place them in their corresponding positions to create a larger image. To input this into the 2D stable diffusion model, we resize the generated image to 512×512 pixels. The entire process is differentiable, allowing the model parameters to be updated through backpropagation.
        
    \subsection{Zero shot 3D Assembly with 2D Diffusion Model} \label{appendix sec: 3d assembly with 2d diffusion}
        In this experiment, we designed a model composed of a two-layer Transformer-Encoder and a Multi-Layer Perceptron (MLP), which is used to predict the rotation angles for each part. We still use trainable embedding as input to represent each part. We also used the SDS loss function here to optimize the model. The prompt we used is "a picture of a 
        
        \noindent colorful chair," because each of our parts is colorful. To achieve a differentiable rendering of 3D point clouds, we utilized the FoV Orthographic Camera from the PyTorch3D library. The model can also update its parameters through backpropagation.

    \subsection{Consider from a training perspective why the supervised method fails on zero-shot scenario.} \label{appendix sec: baseline fail}
        The translation component is supervised by Euclidean loss $\mathcal{L}_t$, which measures the distance between the predicted translation $T_i$ and ground-truth translation $T_i^*$ for each fracture, formulated as:
        \begin{equation}
            \mathcal{L}_t = \sum^N_{i=1} \left | \left | T_i - T_i^* \right | \right | _{2} ^{2}.
        \end{equation}

        For rotation, we employ the Chamfer distance on the rotated point clouds of the parts, defined as:
        \begin{equation}
        \begin{aligned}
            \mathcal{L}_{r} &= \sum_{i=1}^N\left( \sum_{x \in R_i (P_i)} \min _{\substack{y \in R_i^*(P_i)}}\|x-y\|_2^2  \right. \\
            &\left.  + \sum_{x \in R_i^*(P_i)} \min _{\substack{y \in R_i(P_i)}}\|x-y\|_2^2 \right) ,
        \end{aligned}
        \end{equation}
        in which the $R_i (P_i)$ and $R_i^*(P_i)$ represent the rotated fracture points $P_i$ using the estimated rotation $R_i$ and the ground-truth $R^*_i$, respectively.

        To ensure the overall quality of the assembled shape, we incorporate the Chamfer distance (CD) to evaluate the entire shape assembly process for $S$:

        \begin{equation}
            \mathcal{L}_s=\sum_{x \in S} \min _{y \in S^*}\|x-y\|_2^2+\sum_{y \in S^*} \min _{x \in S}\|x-y\|_2^2,
        \end{equation}
        where $S$ is the assembled shape and $S^*$ denotes the ground-truth. 
        The total loss, integrating these components, is thus defined as:
        \begin{equation}
            \mathcal{L} = w_t \mathcal{L}_t + w_r \mathcal{L}_r + w_s \mathcal{L}_s,
        \end{equation}
        % where $w_c$, $w_t$, $w_r$, and $w_s$ denote the weight of different losses, which are empirically determined.
        where $w_t$ = 1.0, $w_r$ = 10, and $w_s$ = 1.0 denote the weight of different losses, which are empirically determined by Zhan et al.~\shortcite{zhan2020generative}.
        Among them, their target for fitting each part ($\mathcal{L}_r$) is much greater than other requirements ($\mathcal{L}_r$ and $\mathcal{L}_r$). This is another reason why their methods cannot be correctly transferred to zero-shot scenarios.

    \section{Reproducibility}

        In this work, we ensure that all results are reproducible. All proofs, code, and data will be provided upon acceptance. The theoretical contributions are clearly stated, with all assumptions and limitations outlined, and appropriate citations for the theoretical tools used. The datasets are publicly available at \url{https://github.com/Wuziyi616/multi_part_assembly}. We have also released the existing codes at \url{https://github.com/Ruiyuan-Zhang/Zero-Shot-Assembly}.

    \clearpage
    \bibliographystyle{named}
    \bibliography{ijcai25}

\end{document}